\pdfoutput=1

\documentclass[11pt]{article}

\usepackage{EMNLP2023}

\usepackage{times}
\usepackage{latexsym}

\usepackage[T1]{fontenc}
\usepackage[utf8]{inputenc}

\usepackage{microtype}

\usepackage{inconsolata}

\usepackage{array}
\usepackage{booktabs}
\usepackage{subcaption}
\usepackage{xspace}
\usepackage{multirow}

\usepackage{nicematrix,tikz}
\usepackage{booktabs}

\newcommand{\metricname}[1]{\textsc{#1}}
\newcommand{\bleu}{\metricname{Bleu}\xspace}
\newcommand{\bleurt}{\metricname{Bleurt}\xspace}
\newcommand{\bleurtqe}{\metricname{BleurtQE}\xspace}
\newcommand{\comet}{\metricname{Comet22}\xspace}
\newcommand{\plaincomet}{\metricname{Comet}\xspace}
\newcommand{\cometqe}{\metricname{CometKiwi}\xspace}
\newcommand{\bicleaner}{\metricname{Bicleaner}\xspace}
\newcommand{\bicleanerAI}{\metricname{Bicleaner~AI}\xspace}
\newcommand{\chrF}{\metricname{ChrF}\xspace}
\newcommand{\none}{None}
\newcommand{\random}{Random}
\newcommand{\lfromto}{$\leftrightarrow$\xspace}
\newcommand{\lto}{$\rightarrow$\xspace}
\newcommand{\paracrawl}{ParaCrawl\xspace}

\newcommand{\starA}{\makebox[0pt]{\hspace{1.25ex}$^\star$}}
\newcommand{\starB}{\makebox[0pt]{\hspace{2.25ex}$^{\star\star}$}}
\newcommand{\starC}{\makebox[0pt]{\hspace{3.25ex}$^{\star\star\star}$}}

\newcolumntype{H}{>{\setbox0=\hbox\bgroup}c<{\egroup}@{}}

\usepackage{numprint}
\npdecimalsign{.}
\nprounddigits{1}
\newcommand{\rp}[1]{\numprint{#1}}

\title{There's no Data Like Better Data: \\ Using QE Metrics for MT Data Filtering}

\author{
  Jan-Thorsten Peter\thanks{\quad Equal contribution.} \\ \textbf{Mara Finkelstein}
  \And David Vilar\footnotemark[1] \\ \textbf{Juraj Juraska} \\
  Google \\
  \texttt{\{jtp,vilar\}@google.com}
  \And Daniel Deutsch \\ \textbf{Markus Freitag}
}

\begin{document}
\maketitle
\begin{abstract}
Quality Estimation (QE), the evaluation of machine translation output without the need of explicit references, has seen big improvements in the last years with the use of neural metrics.
In this paper we analyze the viability of using QE metrics for filtering out bad quality sentence pairs in the training data of neural machine translation systems~(NMT).
While most corpus filtering methods are focused on detecting noisy examples in collections of texts, usually huge amounts of web crawled data, QE models are trained to discriminate more fine-grained quality differences.
We show that by selecting the highest quality sentence pairs in the training data, we can improve translation quality while reducing the training size by half.
We also provide a detailed analysis of the filtering results, which highlights the differences between both approaches.

\end{abstract}

\section{Introduction}
In the times of statistical machine translation, a well-known motto was ``there's no data like more data''.
Experimental results seemed to confirm this, with the performance of the systems steadily improving as more data was made available.
Web crawling has proven to be a valuable source of data for training translation systems, with projects like Common Crawl\footnote{\url{https://commoncrawl.org/}} or \paracrawl \cite{banon-etal-2020-paracrawl} providing numerous parallel sentences with which to train MT systems.
Inevitably, when crawling huge amounts of data, noise will be present.
Taking the web as an example, the quality of available texts varies greatly between websites.
There are sources which reliably produce high-quality text, e.g.\ large circulations newspaper websites usually contain text written and proof-read by professional journalists.
But by its open nature, the web also contains texts of dubious quality (both in style and in content) which may pollute the collected texts.

When considering bilingual data collection, an additional difficulty comes into play, namely the alignment of segments between two or more languages.
Sentence alignment algorithms \cite{gale-church-1993-program,moore-2002-fast,sennrich-volk-2011-iterative,thompson-koehn-2019-vecalign} are bound to make mistakes, resulting in pairing of sentences that are not necessarily translations of each other.
Even if the correspondence between sentences may be correct, the quality of sentences may differ greatly between languages.
While one source may provide high quality text in its original language, the available translations may be of sub-par quality due to a variety of reasons \cite{freitag-etal-2022-natural}.
In addition, given the increased availability (and quality) of machine translation engines, MT output is expected to be part of the crawled data, thus contaminating the training material.

Statistical systems were robust against such type of noise \cite{goutte-etal-2012-impact}.
The maximum likelihood estimators of phrase probabilities (and related models) were based on relative frequencies, with the consequence that noisy translation units, while being available to the system at translation time, had a low chance of being used.
In fact, works on filtering data dealing with statistical systems, e.g.~\citet{johnson-etal-2007-improving} were more concerned with the efficiency of the systems, rather than with quality.

With the advent of neural machine translation, the situation has changed, and the quality of the data has a major impact in the resulting quality of the translation system.
Neural networks have a great ability of memorizing (parts of) the training data \cite{pmlr-v70-arpit17a,NEURIPS2020_1e14bfe2}.
Whereas for phrase-based models the noise was diluted in the abundance of better-quality data, in neural models such outliers may have a critical effect on the output of the system.
Therefore data filtering has become increasingly important, even spawning dedicated shared tasks \cite{koehn-etal-2018-findings,koehn-etal-2019-findings,koehn-etal-2020-findings}.
Research in this area has allowed NMT systems to take advantage of big amounts of training data.
As an example, initial versions of \paracrawl degraded translation quality when adding them unfiltered to the training data of MT systems \cite{junczys-dowmunt-2018-microsofts,schamper-etal-2018-rwth}, whereas nowadays it is one of the main data sources used in WMT for the languages where it is available.

Thanks to these filtering techniques, huge amounts of parallel sentences are available for several languages, in some cases reaching up to hundreds of millions of sentences.
But as noted above, not all texts are of the same quality.
It is only natural to ask the question, once the training data reaches a size which is ``big enough'', if all the available text is useful for training NMT systems, or if the lower quality sentences are hindering the system.
Note that most data filtering systems are focused in detecting noise or problems in the translation (e.g. under- or overtranslation).
In very wide strokes, most systems answer the question ``Is sentence $A$ a translation of sentence $B$?'', without looking (too much) into the quality.

Judging the quality of translations is the focus of the Quality Estimation~(QE) field of machine translation.
It can be considered an extension of machine translation evaluation, where references are not available.
In the last years, the use of neural models has improved the results in this area dramatically.
Would it then be possible to use quality estimation methods for filtering data and improve the quality of a neural machine translation system, which has been trained on already cleaned data?
In other words, can we do a more fine-grained data selection beyond discarding ``obvious'' errors, focusing on selecting the best data that can be found in the training corpus?
We explore these questions in this paper.

Our scientific contributions are:
\begin{itemize}
  \item We show that neural QE metrics are effective methods for data filtering.
  \item We analyze the differences between the sentences filtered by the methods and find out that QE methods are more sensitive toward quality differences, being able to detect bad quality translations or fine grained translation errors (e.g. wrong named entities in a perfectly valid translation).
  \item We show that on the other hand QE filtering cannot account for some actual noise problems.
    Thus a ``traditional'' method for filtering the raw data coming from crawls is still needed as a first step.
\end{itemize}

\section{Related Work}
\label{sec:related_work}

Already in the early days of the popularization of statistical methods for machine translation, the potential of mining data from the web was recognized by \citet{resnik-1999-mining}.
In contrast to other data sources at the time (e.g. Canadian Hansards, European Parliament Proceedings) which consisted largely of clean data, the necessity of including additional ``quality assurance'' steps were recognized in this work.
As pointed out above, statistical systems were robust against noisy input data \cite{goutte-etal-2012-impact}, and as such, the topic of ``corpus filtering'' was mainly focused on selecting subsets of data closer to a given domain \cite{axelrod-etal-2011-domain}.
Nevertheless \citet{taghipour-etal-2011-parallel} shows that statistical systems may also benefit from careful curation of the training data.

The situation changed dramatically with the advent of neural machine translation, as such systems are much more sensitive to noisy input data \cite{khayrallah-koehn-2018-impact}.
A clear reflection of this fact was the creation of a new dedicated shared task in the WMT yearly conference \cite{koehn-etal-2018-findings,koehn-etal-2019-findings,koehn-etal-2020-findings} in the years 2018 to 2020.\footnote{In this year's WMT there is a new related shared task: ``Parallel Data Curation''.}

\citet{junczys-dowmunt-2018-dual} was the best performing system in the first edition of the filtering shared task, using a cross-entropy approach between two translation systems trained on clean data.
In the next edition, the focus was moved towards low resource conditions.
That year \citet{chaudhary-etal-2019-low} presented the best performing system, using a system based on LASER embeddings \cite{schwenk-douze-2017-learning}.

The 2020 edition continued the focus on low-resource languages.
At that evaluation, three were the best performing systems:
\citet{lu-etal-2020-alibaba} and \citet{lo-joanis-2020-improving} both use pre-trained multilingual models as a key component of their filtering systems.
\citet{espla-gomis-etal-2020-bicleaner} used an improved version of \bicleaner, their submission for the 2018 campaign \cite{sanchez-cartagena-etal-2018-prompsits}.
The authors further improved their system \cite{ramirez-sanchez-etal-2020-bifixer}, including an extension using neural models \cite{zaragoza-bernabeu-etal-2022-bicleaner}.
This latest version is considered state-of-the art and it is that which we take as baseline for comparing our method.

Of course, this is just but a very rough overview of the best performing systems in each evaluation campaign.
We refer to the reports of each campaign for a more detailed overview of the methods explored each year.
\citet{bane-etal-2022-comparison} provide one more recent overview of data filtering methods.
In this work, the authors sample 5M sentences from original training data and added 1M noise samples manually.
They show that a two-stage approach can be beneficial for improving the quality of a translation system.

All of these methods have a common focus on detecting the type of noise that may originate from crawled data.
This type of noise has been analyzed in \citet{khayrallah-koehn-2018-impact}, and \citet{herold-etal-2022-detecting} build on their work and carry out a comparison of the efficiency of different filtering methods on various types of noise.
\citet{kreutzer-etal-2022-quality} also provide an extensive analysis on the noise present in several widely used corpora.

Most similar to our work \citet{carpuat-etal-2017-detecting} start with already clean data and analyze the effect of semantic divergence on translation quality.
They are able to effectively select a subset of the training data and improve translation quality measured in \bleu.
\citet{bernier-colborne-lo-2019-nrc} use YiSi-2, also a quality estimation metric as a component in their corpus filtering system for the WMT~2019 shared task.
\citet{lo-simard-2019-fully} extend this idea by including \textsc{BERT} (word) alignments in the YiSi pipeline.
We follow a conceptionally similar approach to these papers, using state-of-the-art QE metrics and provide a more in-depth comparison to other corpus filtering methods more oriented towards noise detection.

Quality estimation is again its own area of research, with dedicated shared tasks, e.g.\ \cite{zerva-etal-2022-findings}, that measure how well metrics can predict word- and sentence-level quality scores.
In contrast to traditional MT evaluation, QE aims to assess the quality of the output texts without the use of a reference translations.
The most successful QE metrics learn to jointly predict word- and sentence-level scores, like \cometqe \cite{kepler-etal-2019-openkiwi,rei-etal-2022-cometkiwi}.
Another possibility is to modify the input to a learned reference-based metrics like \bleurt \cite{sellam-etal-2020-bleurt} or \plaincomet \cite{rei-etal-2020-comet} to use the source segment instead of a reference translation to predict sentence-level quality scores \cite{rei-etal-2021-references}.
We follow the latter approach and train a QE version of BLEURT that predicts sentence-level quality scores (see Section~\ref{sec:bleurtqe}) that are used for data filtering.

\section{From \bleurt to \bleurtqe}
\label{sec:bleurtqe}

The QE metric that we propose for data filtering is a learned MT evaluation metric that is based on a \bleurt-style architecture \cite{sellam-etal-2020-bleurt}.
\bleurt is a reference-based regression metric that is trained to predict a quality score for a hypothesis translation given a reference.
The hypothesis and reference are concatenated together with a special token in between, then fed as input to the metric, which predicts a floating point quality score.

Our QE metric is a modification of the original model.
To make it a QE metric, we pass the source segment as input to the metric instead of the reference.
Then, we follow the winning submission to the WMT'22 Metrics Shared Task \cite{freitag-etal-2022-results}, MetricX, and use a modified version of the mT5 encoder-decoder language model \cite{xue-etal-2021-mt5} as our network architecture.
Not that these is a multilingual model, so the same system can be used for a variety of languages.
The source and hypothesis are passed as input to the encoder, and an arbitrary logit from the first step of the decoder is trained to predict the hypothesis quality score.

The QE metric is trained on the direct assessment quality judgments that were collected as part of the WMT Metrics Shared Task from 2015-2020 \cite{bojar-etal-2015-findings,bojar-etal-2016-findings,bojar-etal-2017-findings,specia-etal-2018-findings,specia-etal-2020-findings-wmt,fonseca-etal-2019-findings} for all available language pairs.
To (meta-)evaluate the metric we measure its correlation with ground-truth translation quality ratings using the benchmark MQM dataset from WMT'22 \cite{zerva-etal-2022-findings} that includes 3 language pairs: en-de, zh-en, and en-ru.
Since our metric is used to score individual segments and not systems, we report the segment-level correlation between our metrics' scores and the gold MQM scores using Pearson's~$r$ and Kendall's~$\tau$, shown in Table~\ref{tab:qe_meta_eval}.
The correlations are competitive to the top QE submissions to the WMT'22 Metrics Shared Task.

A (more refined) version of this metric has been submitted to this year's QE shared task \cite{juraska2023metricx}, and has been open sourced.
We refer the reader to the system description for a more fine-grained discussion of the details of the metric.

\begin{table}
    \centering
    \small
    \setlength{\tabcolsep}{5pt} %
    \begin{tabular}{lcccccc}
        \toprule
        & \multicolumn{2}{c}{\bf en-de} & \multicolumn{2}{c}{\bf en-ru} & \multicolumn{2}{c}{\bf zh-en} \\
        \cmidrule(lr){2-3} \cmidrule(lr){4-5} \cmidrule(lr){6-7}
        \bf Metric & $r$ & $\tau$ & $r$ & $\tau$ & $r$ & $\tau$ \\
        \midrule
        UniTE-src & 0.40 & 0.29 & 0.39 & 0.34 & 0.40 & 0.43 \\
        \textsc{CometKiwi} & 0.43 & 0.29 & 0.39 & 0.36 & 0.51 & 0.36 \\
        \bleurtqe & 0.38 & 0.29 & 0.41 & 0.39 & 0.38 & 0.35 \\
        \bottomrule
    \end{tabular}
    \caption{Segment-level Pearson's $r$ and Kendall's $\tau$ on the WMT'22 MQM ratings for our QE metric, \bleurtqe and the top-performing metrics in the WMT'22 Metrics Shared Task, \textsc{CometKiwi} \cite{rei-etal-2022-cometkiwi}, UniTE-src \cite{wan-etal-2022-unite}. %
    }
    \label{tab:qe_meta_eval}
\end{table}

\section{Experiments}
\label{sec:experiments}

We report experiments on three language pairs: English \lfromto German, Japanese \lfromto English and Chinese \lfromto English.
Our starting point is the full training data as provided by the WMT evaluation campaign.
Corpus sizes can be found in Table~\ref{tab:compare-filtering-sizes}.
As can be seen in that table, we are working on a medium-to-large data condition, with the smallest language pair already having over 30M sentence pairs.

One thing to note is that these datasets have already undergone a cleaning process by the WMT organizers.
I.e.\ a system trained on the entirety of this data is already able to obtain very good performance.
In fact, many of the systems participating in the WMT evaluations take the available data as-is.

For each language pair we will consider different ways to reduce the size to 50\% of their original size.
This value was chosen in preliminary experiments on the English to German data, and it is comparable to previous work \cite{bane-etal-2022-comparison}.
Fixing the target size beforehand also allows a fair comparison between all the methods.

\subsection{Filtering Approaches}

We will consider three different filtering approaches for our experiments.

\subsubsection{Random Selection}

The most straightforward method to reduce the size of the training data is to just randomly select the desired amount of sentence pairs.
We do not expect this method to perform well, but it constitutes the most direct baseline for data size reduction.

\subsubsection{\bicleaner}

As a representative for the ``noise-detection'' corpus filtering methods we chose to use \bicleanerAI.\footnote{\url{https://github.com/bitextor/bicleaner-ai}}
This tool is an extension of the previous \bicleaner tool.
The underlying method is based on a classifier that predicts if a sentence is a translation of another.
\bicleanerAI substitutes the original classifier, based on handcrafted rules and extremely randomized trees, with a neural classifier based on XLM-RoBERTa.
\citet{zaragoza-bernabeu-etal-2022-bicleaner} provide a detailed description of the tool and present an extensive experimental comparison showing state of the art results for filtering \paracrawl.

It is also worth noting that \bicleaner is part of the pre-processing pipeline for generating the \paracrawl dataset.

\subsubsection{Quality Estimation for Filtering}

For testing the performance of QE metrics for filtering we use two state-of-the-art metrics, \cometqe\footnote{\url{https://unbabel.github.io/OpenKiwi}} \cite{kepler-etal-2019-openkiwi} and \bleurtqe\footnote{The tool will be open-sourced with the publication of the shared task system description.} as described in Section~\ref{sec:bleurtqe}.
For each sentence pair in the training data, we compute the QE score for the translation from English into the foreign language.
We use these scores for filtering for both translation directions, i.e. the resulting parallel data is the same for English \lto Foreign than from Foreign \lto English.
We are aware that this may introduce a certain bias, as the performance of the QE metrics is not symmetrical.
However scoring the full training data is a costly operation as we have to run big neural models on tens or hundreds of millions of sentence pairs.
We still expect to see improvements even when using the wrong direction for data filtering.
The only exception may be the backtranslated portion of the Chinese \lfromto English dataset: As the starting data is Chinese, the filtering method may miss low quality backtranslations produced by an automatic system.

\subsection{Experimental Setup}

For all the filtering methods (except random selection), we compute the score of each sentence pair, and then select a threshold as to keep 50\% of the original data.
We then train an NMT system from scratch using the resulting training data sets.

Our translation system is a transformer-based encoder-decoder model based on PaxML\footnote{\url{https://github.com/google/paxml}}, very similar to most of the systems participating in the WMT evaluation campaign.
It consists of 6 encoder and 6 decoder layers, a model dimension of 1024, hidden dimension of 8192 and 16 attention heads.
GELUs with gated activation are used as activation functions.
We use a 32k shared vocabulary for each language pair, and limit the maximal sentence length to 128 tokens.
The model has a total of 551M parameters.

We removed all sentences which have more than 128 tokens, but did not perform any other filtering or preprocessing of the data.
All models are trained until they converged and we selected the checkpoint with the best \bleurt score on the WMT 2022 test sets.

In the discussion of the results we focus on the evaluation using \comet.
Traditional metrics like \bleu and \chrF are consistently outperformed by neural metrics in the WMT metrics shared task \cite{freitag-etal-2022-results}, thus we favor the use of such new metrics.
We chose \comet over \bleurt in order to avoid overfitting on this last metric, as our proposed \bleurtqe model is based on it, and it also guides the checkpoint selection.
Nevertheless, \bleurt, \bleu and \chrF scores are given in Appendix~\ref{sec:other-metrics} and confirm the trends reported here.

\subsection{Test Data}

In order to test on a variety of domains we use test sets from the WMT and IWSLT evaluation campaigns.
We use the WMT 2019 (where available) consisting of news data, and the WMT 2022 and WMT2023 test sets, which are composed of a mix of different domains each.
Additionally we experiment on the IWSLT'21 test set, sourced from TED talks \cite{anastasopoulos-etal-2021-findings}, and the IWSLT'23 dev set\footnote{We use the dev set for IWSLT'23, since the test set is currently not publicly available.}, which is based on ACL talks presentations \cite{agrawal-etal-2023-findings}.

Following the training data settings, we also filtered the test sentences longer than 128 tokens.
As the WMT 2023 test set includes paragraph level evaluation, its size is reduced for en \lto de from 557 segments to 404 and for de \lto en from 549 to 468.
All other test sets are barely affected (see Table~\ref{tab:testset_filtering} in Appendix~\ref{sec:test_data_stats}).

\subsection{Experimental Results}

\begin{table*}
  \newcolumntype{e}{>{\raggedright\arraybackslash}p{5cm}}
  \newlength{\betweenTabs}
  \setlength{\betweenTabs}{0.5cm}

  \begin{subtable}{\textwidth}
	  \subcaption{\comet scores for en \lfromto de experiments.}
    \label{tab:endecomet}
  \begin{center}
    \begin{NiceTabular}{llrrrrr}
      \toprule
                             & Filter     & {WMT'22 (dev)}      & {WMT'19}            & {WMT'23}            & {IWSLT'21}          & {IWSLT'23}          \\
      \midrule
 \multirow{5}{*}{en \lto de} & \random    & \rp{84.0}           & \rp{85.50}          & \rp{80.80}          & \rp{82.80}          & \rp{84.20}          \\
	    & \none      & \rp{86.18}          & \rp{85.99}          & \rp{81.81}          & \rp{83.22}          & \rp{84.53}          \\
                             & \bicleaner & \rp{86.67}\starA          & \rp{86.42}\starA          & \rp{82.20}          & \rp{83.29}          & \rp{84.85}          \\
			     & \cometqe   & \rp{86.92}\starB          & \textbf{\rp{86.66}}\starB & \rp{82.93}\starA          & \rp{83.57}\starA          & \rp{84.81}          \\
                             & \bleurtqe  & \textbf{\rp{87.19}}\starC & \textbf{\rp{86.65}}\starB & \textbf{\rp{83.49}}\starB & \textbf{\rp{83.91}}\starB & \textbf{\rp{85.08}}\starA \\
      \midrule
 \multirow{5}{*}{de \lto en} & \random    & \rp{84.2}           & \rp{84.30}          & \rp{83.50}          & \rp{84.10}          & \rp{87.20}          \\
	    & \none      & \rp{84.50}          & \rp{84.60}          & \rp{83.32}          & \rp{84.22}          & \textbf{\rp{87.44}} \\
                             & \bicleaner & \rp{84.23}\starA          & \rp{84.77}          & \rp{84.02}\starA          & \rp{84.09}          & \rp{87.28}          \\
                             & \cometqe   & \rp{84.64}\starA          & \rp{85.07}\starA          & \textbf{\rp{84.59}}\starB & \textbf{\rp{84.39}}\starA & \rp{87.32}          \\
                             & \bleurtqe  & \textbf{\rp{84.78}}\starB & \textbf{\rp{85.18}}\starA & \textbf{\rp{84.64}}\starB & \textbf{\rp{84.35}}\starA & \rp{87.29}          \\
      \bottomrule
      \CodeAfter
\tikz \draw [dashed, shorten < = 4pt, shorten > = 4pt] (3-|2) -- (3-|8) ;
\tikz \draw [dashed, shorten < = 4pt, shorten > = 4pt] (8-|2) -- (8-|8) ;
    \end{NiceTabular}
  \end{center}
  \end{subtable}
  \vspace*{\betweenTabs}

  \begin{subtable}{\textwidth}
  \begin{center}
  \subcaption{\comet scores for en \lfromto ja experiments.}
    \label{tab:enjacomet}
    \begin{NiceTabular}{llrrr}
      \toprule
                                & Filter     & WMT'22 (dev)        & WMT'23              & IWSLT'23            \\
      \midrule
    \multirow{4}{*}{en \lto ja} & \random    & \rp{84.50}          & \rp{80.70}          & \rp{85.80}          \\
	    & \none      & \rp{85.63}          & \rp{82.34}          & \rp{86.89}          \\
                                & \bicleaner & \rp{85.95}\starA          & \rp{83.21}\starA          & \rp{87.21}          \\
                                & \cometqe   & \rp{86.56}\starB          & \rp{83.65}\starB          & \textbf{\rp{87.92}}\starA \\
                                & \bleurtqe  & \textbf{\rp{86.99}}\starC & \textbf{\rp{83.99}}\starC & \rp{87.42}          \\
      \midrule
    \multirow{4}{*}{ja \lto en} & \random    & \rp{75.90}          & \rp{75.00}          & \rp{84.60}          \\
	    & \none      & \rp{77.63}          & \rp{75.89}          & \textbf{\rp{85.47}}\starA \\
                                & \bicleaner & \rp{78.13}\starA          & \rp{77.42}\starA          & \rp{85.02}          \\
                                & \cometqe   & \rp{78.69}\starB          & \rp{77.97}\starB          & \rp{84.95}          \\
                                & \bleurtqe  & \textbf{\rp{78.98}}\starC & \textbf{\rp{78.21}}\starB & \rp{85.11}          \\
      \bottomrule
      \CodeAfter
\tikz \draw [dashed, shorten < = 4pt, shorten > = 4pt] (3-|2) -- (3-|6) ;
\tikz \draw [dashed, shorten < = 4pt, shorten > = 4pt] (8-|2) -- (8-|6) ;
    \end{NiceTabular}
  \end{center}
  \end{subtable}
  \vspace*{\betweenTabs}

  \begin{subtable}{\textwidth}
  \begin{center}
  \subcaption{\comet scores for en \lfromto zh experiments.}
    \label{tab:enzhcomet}
    \begin{NiceTabular}{llrrrr}
      \toprule
                                  & Filter     & WMT'22 (dev)        & WMT'19              & WMT'23              & IWSLT'23            \\
      \midrule
      \multirow{4}{*}{en \lto zh} & \random    & \rp{80.20}          & \rp{77.20}          & \rp{79.30}          & \rp{82.10}          \\
	    & \none      & \rp{81.24}          & \rp{77.56}          & \rp{79.71}          & \textbf{\rp{84.24}}\starA \\
                                  & \bicleaner & \rp{81.70}\starA          & \rp{78.39}\starA          & \rp{80.30}\starA          & \rp{83.31}          \\
                                  & \cometqe   & \rp{83.00}\starB          & \textbf{\rp{80.14}}\starB & \textbf{\rp{82.51}}\starC & \rp{84.09}\starA          \\
                                  & \bleurtqe  & \textbf{\rp{83.37}}\starC & \rp{79.93}\starB          & \rp{82.21}\starB          & \rp{84.11}\starA          \\
      \midrule
\multirow{4}{*}{zh \lto en}       & \random    & \rp{72.20}          & \rp{78.10}          & \rp{74.20}          & \rp{84.10}          \\
	    & \none      & \rp{72.77}          & \rp{78.32}          & \rp{74.72}          & \textbf{\rp{84.90}}\starA \\
                                  & \bicleaner & \rp{74.82}\starA          & \rp{79.57}\starA          & \rp{75.66}\starA          & \rp{84.20}          \\
                                  & \cometqe   & \rp{75.24}\starB          & \textbf{\rp{79.97}}\starB & \textbf{\rp{75.96}}\starB & \rp{84.45}          \\
                                  & \bleurtqe  & \textbf{\rp{75.36}}\starB & \rp{79.90}\starB          & \textbf{\rp{75.96}}\starB & \rp{84.78}\starA          \\
      \bottomrule
      \CodeAfter
\tikz \draw [dashed, shorten < = 4pt, shorten > = 4pt] (3-|2) -- (3-|7) ;
\tikz \draw [dashed, shorten < = 4pt, shorten > = 4pt] (8-|2) -- (8-|7) ;
    \end{NiceTabular}
  \end{center}
  \end{subtable}

    \caption{\comet scores all experiments.
    For each language direction, systems marked with stars are statistically significantly better than systems with fewer stars (pairwise permutation test \cite{koehn-2004-statistical} with p=0.05).
    "Random" was excluded from the significance computation.
    }
\end{table*}

Translation results for the English \lfromto German language pair are shown in Table~\ref{tab:endecomet}.
For en \lto de we can see that randomly selecting data hurts performance by 1 point on the WMT23 test set.
Using each of the other filtering methods we are able to improve performance over using the full training dataset.
For \bicleaner the improvement is rather modest, around 0.4 points for most test corpora.
Note however that \bicleaner was already applied to the \paracrawl dataset, which constitutes a big portion of the available training data for this language pair.
As such it is understandable, or even expected, that translation quality is not improved by applying it again.
The QE metrics perform similarly to each other, with a slight advantage of \bleurtqe over \cometqe.
Using \bleurtqe we are able to achieve an improvement of up to 1.7 points on the WMT23 test set.

The results for de \lto en majorly confirm the previous observations.
The best results are again achieved in this case on the WMT'23 data, with an improvement of 1.3 points achieved by both QE methods.
For the WMT'22, IWSLT'21 and IWSLT'23 test sets, the translation performance basically stays constant for all filtering methods.

Results on English \lfromto Japanese, shown in Table~\ref{tab:enjacomet} also show similar trends.
In this case the biggest improvement comprises 2.3 points on the WMT'23 test set\footnote{A slightly bigger improvement of 2.4 is obtained for WMT'22, but we skip this as we used this corpus to choose the best checkpoint during training.}, obtained by \bleurtqe.
However for the ja \lto en translation direction we find an outlier, where no filtering achieves improvements over the baseline on the IWSLT'23 data.

Lastly, Table~\ref{tab:enzhcomet} shows the results for the Chinese \lfromto English language pair.
Again we can confirm the same trends as for the other two language pairs.
The QE metrics are able to improve up to 2.8 points for the WMT'23 test set.
The IWSLT'23 dataset again fails to achieve improvements, and in this case \bicleaner deteriorates translation quality, while the QE metrics are able to keep the performance.

Overall, we see that the QE metrics are effective in improving translation quality while retaining just half of the training data.
The improvements can rage up to more than 2 \comet points, depending on language pair and test set.
For IWSLT'23, having a more specialized technical domain, the QE metrics are not able to improve quality, for several language directions.
But except for the case of English \lto Japanese, they also do not hurt performance.
Additional results differentiating between the single domains of the WMT'22 and WMT'23 corpora can be found in Appendix~\ref{sec:wmt-domain-specific}.
In Appendix~\ref{sec:other-metrics} we also report the results of the same experiments using \bleu, \chrF and \bleurt.
These metrics confirm the observations presented in this section.

\citet{koehn-etal-2020-findings} mentions that on average metrics that select shorter sentences performed better on Parallel Corpus Filtering.
Contrary to that we observed that dropping 50\% of the data with the proposed method led to on average slightly longer sentences e.g. from 14.4 to 15.5 words per sentence for \bleurtqe on English \lfromto German.

\section{Analysis}
\label{sec:analysis}

In this section we provide an in-depth analysis of the differences between the \bleurtqe-based and the \bicleaner filtering methods.
Table~\ref{tab:compare-filtering-sizes} shows the amount of sentence pairs that are kept by both methods, which is roughly two thirds of the filtered sentences for all language pairs.
Thus, it is clear that both methods do indeed perform quite different filtering.
We will first report on manual inspection of the most striking divergences between both methods.
In Section~\ref{sec:noise-analysis} we will then provide a more quantitative analysis of the behaviour of the methods using synthetic data.

\begin{table}
  \begin{center}
    \begin{tabular}{lrrrr}
      \toprule
	    Language pair   & Full & Filtered & Common \\
      \midrule
	    de \lfromto en  & 292.8M & 146M     & 105M   \\
	    en \lfromto zh  & 55.2M  & 27.6M    & 17.5M  \\
	    en \lfromto ja  & 33.9M  & 16.9M    & 10.4M  \\
      \bottomrule
    \end{tabular}
  \end{center}

    \caption{
  Amount of sentences before filtering, after filtering, i.e.\ 50\% of the original corpus size, and number of sentences kept by both \bleurtqe and \bicleaner.
  All language directions include \paracrawl data.
  English \lfromto Chinese includes around 19.7M backtranslated Chinese sentences, as provided by the WMT organizers.
  }
    \label{tab:compare-filtering-sizes}
\end{table}

\subsection{Human Inspection}
\label{sec:human-inspection}

We will now analyze the difference in the filtering methods by looking into the sentences that are selected by each method.
To this end, we select the sentences where one method filters it but the other does not.
In addition we use automatic clustering methods in the spirit of \cite{aharoni-goldberg-2020-unsupervised} in order to get insights about topic distribution.
We limit our analysis to the German--English language pair\footnote{None of the authors are speakers of Japanese or Chinese.}, but as the methods are largely language independent, we feel confident that our findings will generalize to the other language pairs.
Also, due to the fuzzy and partially subjective nature of this investigation, we are unable to provide exact statistics about each kind of effect.\footnote{If we were able to develop such statistics in an automatic way, we would be able to improve the filtering methods by including the same approaches!}

We have encountered the following major differences in the working of the methods.
For each of these categories it is easy to find an abundance of examples (easily in the thousands) in the filtered data.

\begin{table*}
  \newcolumntype{e}{>{\raggedright\arraybackslash}p{5cm}}
  \setlength{\betweenTabs}{0.5cm}

  {\small
  \begin{subtable}{\textwidth}
    \subcaption{
      Single Entity Mistranslations.
      Templated texts where the specific instantiation is different in both languages.
      \bleurtqe filters out these examples, while \bicleaner keeps them.
    }
    \label{tab:se-mistranslations}
    \begin{center}
      \begin{tabular}{eee}
        \toprule
        English & German & Comments \\
        \midrule
        Flights from Tallinn to Stockholm &
        Flüge ab Tallinn nach Friedrichshafen &
        ``Stockholm''' changed to ``Friedrichshafen''.
        \\ \midrule
        Total EU spending in Germany – € 11.013 billion &
        Gesamtzuschüsse der EU in den Niederlanden: 2,359 Milliarden EUR &
        Land and amount changed.
        \\ \midrule
        Documents that we receive from a manufacturer of a Redball Electrical 565 can be divided into several groups. &
        Dokumente, die wir vom Produzenten des Geräts Trevi AVX 565 erhalten, können wir in mehrere Gruppen teilen. &
        Product code changed. \\
        \bottomrule
      \end{tabular}
    \end{center}
  \end{subtable}
  \vspace*{\betweenTabs}

  \begin{subtable}{\textwidth}
    \subcaption{
      Examples of low quality sentences in at least one of the languages.
      \bleurtqe filters out these examples, while \bicleaner keeps them.
  }
    \label{tab:low-quality-trans}
    \begin{center}
      \begin{tabular}{eee}
        \toprule
        English & German & Comments \\
        \midrule
        We are both, we have own factory which can ensure sculpture quality and best price and have a profession team to provide you best service. &
        Wir sind beide, wir haben eigene Fabrik, die Skulpturqualität und besten Preis sichern kann und ein Berufsteam haben, um Ihnen besten Service zur Verfügung zu stellen. &
        Unnatural language on both sides.
        \\ \midrule
        We honor do not track signals and do not track, plant cookies, or use advertising when a Do Not Track (DNT) browser mechanism is in place. &
        Wir achten darauf, dass Sie keine Signale verfolgen und keine Cookies verfolgen oder Cookies verwenden, wenn Sie einen DNT-Browser-Mechanismus (Do not Track) verwenden. &
        Unnatural language on both sides.
        \\ \midrule
        It really is fast, easy, free and additionally to attempt. &
        Es ist schnell, Schnell, gratis und am besten von allen zu try. &
        Incorrect sentences in both languages.
         \\
        \bottomrule
      \end{tabular}
    \end{center}
  \end{subtable}
  \vspace*{\betweenTabs}

  \begin{subtable}{\textwidth}
    \subcaption{
    Examples of wrong sentence alignment, although the sentences are related to each other.
    \bleurtqe filters out these examples, while \bicleaner keeps them.
  }
    \label{tab:wrong-alignment}
    \begin{center}
      \begin{tabular}{eee}
        \toprule
        English & German & Translated German \\
        \midrule
        Could you help me? Help to improve my English and French language &
        Ja, ich möchte gern mein Deutsch mit Dir verbessern. &
        Yes, I want to improve my German with you.
        \\ \midrule
        We, therefore, guarantee that you will get daily updates on office spaces to rent in Hong Kong. &
        Wir können Ihnen deshalb versichern, dass Sie bei uns täglich einen aktuellen Überblick über den österreichischen Markt erhalten. &
        We can guarantee that you will get an up-to-date daily overview about the Austrian market.
        \\ \midrule
        This implies that the law is either repealed or not enforced. &
        Darüber hinaus wird sichergestellt, dass bestehende Gesetze nicht dupliziert oder konterkariert werden. &
        In addition, it is ensured that existing laws are not duplicated or counteracted. \\

        \bottomrule
      \end{tabular}
    \end{center}
  \end{subtable}
  \vspace*{\betweenTabs}

  \begin{subtable}{\textwidth}
    \subcaption{Examples of sentence pairs originating from the Bible.
    \bleurtqe filters them out, probably due to archaic language, while \bicleaner keeps them.
  }
    \label{tab:domain-specific}
    \begin{center}
      \begin{tabular}{p{7cm}p{7cm}H}
        \toprule
        English & German & Comments \\
        \midrule
        19 Behold, my belly is as wine which hath no vent; it is ready to burst like new bottles. &
        19 Siehe, mein Bauch ist wie der Most, der zugestopfet ist, der die neuen Fässer zerreißet. &
        \\ \midrule
        7:16 Those who went in, went in male and female of all flesh, as God commanded him; and Yahweh shut him in. &
        7:16 und das waren Männlein und Fräulein von allerlei Fleisch und gingen hinein, wie denn Gott ihm geboten hatte.
        \\
        \bottomrule
      \end{tabular}
    \end{center}
  \end{subtable}
  }

  \caption{
    Example sentences where the filtering methods diverge.
  }
\end{table*}

\newcommand{\filterdiff}[1]{\textbf{#1}}

\filterdiff{Single Entity Mistranslations}
When looking into the parallel data available for training, one can find a big amount of ``templated texts'', i.e. sentences that have a common structure, but that differ in one or few components, frequently named entities or numbers.
Some examples can be found in Table~\ref{tab:se-mistranslations}.
The first entry in this table is a typical example.
In the travel domain, there is a big amount of sentences of the form ``Flights from \textit{cityA} to \textit{cityB}'', ``Hotels in \textit{city}'' or similar formulations.
One frequent source of sentence alignment errors originates from sentences that follow the same template, but have different instantiations.
Although the travel domain is one of the biggest representative of these type of sentences, it is by no mean the only one, as the other examples in Table~\ref{tab:se-mistranslations} show, including the financial and the technical domain.

In these type of sentences, the QE metric seems to be more sensitive to alignment errors.
All the sentences shown in Table~\ref{tab:se-mistranslations} (and many others) are selected by \bicleaner, while \bleurtqe discards them.

\filterdiff{Low Quality Translations}
In this category we include training examples where one or both sides are of low quality.
Examples can be found in Table~\ref{tab:low-quality-trans}.
We can see that the language quality of the examples is borderline at best.
Strictly speaking, the translations are ``correct'' in the sense that they preserve the structure of the sentence.
As such \bicleaner gives them a relative high score and are kept in the training corpus.
\bleurtqe, on the other hand, is explicitly trained to flag such erroneous sentences (as they might very well originate from MT engines), and thus these examples are filtered out.

\filterdiff{Bad Related Sentence Alignments}
As pointed out above, sentence alignment is also an automatic process.
While both methods perform quite well when detecting clearly bad aligned sentences (see Section~\ref{sec:noise-analysis}), we found that there are cases where the source and target sides are related and the \bicleaner system seems to get confused by this proximity.
Some examples are given in Table~\ref{tab:wrong-alignment}.
It can be seen that in all three examples the German side is clearly related to the English text, with probably a overlap big enough to get an acceptable score from the translation system underlying \bicleaner.
Again, as \bleurtqe is trained to distinguish fine-grained differences between translations, it is more robust against this kind of problems.

\filterdiff{Religious Texts}
One shortcoming we found for the \bleurtqe method is that many sentences originating from the Bible corpus (or similar religious texts) are filtered out, while \bicleaner keeps them.
Some examples are given in Table~\ref{tab:domain-specific}.
This is probably due to the language being archaic, very different to the type of sentences \bleurtqe has been trained on.
Such style would be heavily penalized in an evaluation, as a more modern language would be preferred.

\subsection{Noise}
\label{sec:noise-analysis}

In the previous section we saw several examples where \bleurtqe outperforms \bicleaner for data selection.
However we should not forget that \bicleaner was developed with a (related but) different goal, namely the cleaning of raw data.
In fact, our starting datasets, as made available for the WMT evaluation have already undergone a cleaning process, and are already at a pretty high quality level.

If we were dealing with crawled data directly, we would need to address different phenomena.
In this section we study how the filtering methods perform when dealing with the typical noise found on crawled data.
We follow \citet{herold-etal-2022-detecting} for the categorization of different noise types, which in turn is based on \citet{khayrallah-koehn-2018-impact}.
We create synthetic data for the English \lto German translation direction containing the following noise categories:
\begin{description}
  \item[Misaligned Sentences] created by shuffling the target side of the corpus.
  \item[Misordered Words] created by reordering the words in either the source or the target sentences.
  \item[Wrong Language] created by taking parallel sentences corresponding to another language pair.
  \item[Untranslated] created by copying one sentence into the other direction i.e. each sentence pair in the corpus has a copy of the source sentence as a ``target'' sentence (or the reverse direction).
  \item[Over/Undertranslation] created by truncating either the source or the target side.
\end{description}
We refer the reader to \citet{herold-etal-2022-detecting} for a more detailed description and justification of these categories.
We omitted the ``Short Segments'', ``Raw Crawled Data'' and ``Synthetic Translations'' categories, as it was not clear how to define the correct filtering strategy in those cases.

For each of the studied categories, we generated 200K synthetic noise examples by randomly selecting a subset of the training data.
For these experiments we re-tuned the threshold for each method by computing the scores for the original sentences and the noise examples, and computing the median.
In this way, we filter exactly half of the data and a perfect system would be able to completely separate the original examples from the noisy ones.

Results can be found in Table~\ref{tab:noise-detection}.
It can be seen that for most categories \bicleaner clearly outperforms \bleurtqe.
This is specially the case for the ``Wrong Language'' and ``Untranslated'' categories, where \bicleaner can detect all the noisy examples.
In fact, one of the practical advantages of \bleurtqe is at the same time one of its weaknesses.
As its backbone model is a multilingual model, it is able to handle a wide number of languages, but it does not have a way to differentiate between them.

For ``Misordered Words'' we find an interesting asymmetry.
\bleurtqe is much stronger in detecting problems when the target side is reordered, undoubtedly due to this being the ``natural'' direction for which it was trained.
\bicleaner also shows this behavior, with its target side performance being superior to that of \bleurtqe, but inferior in the opposite direction.
\bicleaner is also clearly better at detecting Over- and Undertranslations.

\begin{table}
  \begin{center}
    \small
    \begin{tabular}{lrrH}
      \toprule
      Noise type             & \bleurtqe     & \bicleaner    & \cometqe      \\ \midrule
      Misaligned Sentences   & 8.1           & \textbf{5.6}  & 9.4           \\
      Misordered Words (src) & \textbf{24.3} & 39.1          & \textbf{18.1} \\
      Misordered Words (tgt) & 10.3          & \textbf{6.3}  & 10.7          \\
      Wrong Language         & 43.8          & \textbf{0.0}  &               \\
      Untranslated (src)     & 47.6          & \textbf{0.0}  & 14.7          \\
      Untranslated (tgt)     & 63.8          & \textbf{0.0}  & 62.0          \\
      Overtranslation        & 35.2          & \textbf{13.9} & 30.3          \\
      Undertranslation       & 13.4          & \textbf{7.5}  & 12.5          \\
      \bottomrule
    \end{tabular}
  \end{center}

  \caption{
    Percentage of sentences being kept as valid for each of the synthetic noise categories.
    0\% means that all noisy sentences have been filtered out, i.e. perfect performance.
  }
  \label{tab:noise-detection}
\end{table}

\subsection{Combination of Filtering Methods}

Since \bleurtqe and \bicleaner based filtering both improve translation quality, and they filter different sentences, it is only natural to try to combine both.
As can be seen in Table~\ref{tab:compare-filtering-sizes}, the amount of available data dropped to roughly one third when combining both methods.
The result only slightly degraded in the English to German direction compared to just using \bleurtqe (0.3 \bleurt points on WMT'22), but degraded more in the German to English direction (0.6 \bleurt points on WMT'22).
Combining the two methods is thus too aggressive with our setup, and hurts translation performance.
Adapting the thresholds for the combination, may result in better performance.
Note however that \paracrawl already used \bicleaner in its pipeline \cite{espla-etal-2019-paracrawl}, thus we have already implicitly been using a combination of both methods.

\section{Conclusions}

In this paper we have shown that filtering data using QE metrics is an effective way of improving translation quality.
In contrast to ``traditional'' data filtering methods that focus on detecting noise in the data, QE methods focus on selecting the best translation examples.
Analyzing the differences between the two different methods, we see that QE metrics are not as effective at detecting certain types of noise, e.g.\ untranslated sentences, but are much better at identifying more fine grained problems in the data, like small translation errors or grammatical mistakes.
Therefore, when starting with already cleaned data, we can obtain a boost in performance by focusing the NMT system training on the best sentences.

Our results show that the improvements obtained generalize across different domains, as measured by a variety of metrics.
Even for more distant domains, like the ACL Talks of of the IWSLT'23 corpus, the performance of the systems remains largely constant.
QE estimation is a very active field of research.
Using this approach, the improvements obtained in this area can have a direct impact on improving the quality of NMT systems.

\section*{Limitations}

Better results could have been obtained by tuning the threshold for each method individually, but this would also increase the computational cost massively.

A more in-depth comparison could be carried out starting from the raw web-crawled data.
However in this study we chose to start from conditions similar to what most participants in the WMT evaluation use.

\section*{Ethics Statement}
\bleurtqe and \cometqe scoring all the training data is computationally expensive, and may be a limiting factor of the method for small institutions.

\bibliography{custom,anthology}
\bibliographystyle{acl_natbib}

\appendix

\section*{Appendices}

\section{Test Data Statistics}
\label{sec:test_data_stats}

Table~\ref{tab:testset_filtering} shows statistics of the test data sets after filtering the sentences with a length of over 128 tokens.

\begin{table}[ht]
  \begin{center}
    \begin{tabular}{llrr}
      \toprule
                              & test set   & lines & filtered \\
      \midrule
      \multirow{6}{*}{WMT'22} & en \lto de & 2037  & 2036     \\
                              & de \lto en & 1984  & 1981     \\
                              & en \lto ja & 2037  & 2037     \\
                              & ja \lto en & 2008  & 2007     \\
                              & en \lto zh & 2037  & 2037     \\
                              & zh \lto en & 1875  & 1849     \\
      \midrule
      \multirow{6}{*}{WMT'23} & en \lto de & 557   & 404      \\
                              & de \lto en & 549   & 468      \\
                              & en \lto ja & 2074  & 2073     \\
                              & ja \lto en & 1992  & 1988     \\
                              & en \lto zh & 2074  & 2073     \\
                              & zh \lto en & 1976  & 1948     \\
      \bottomrule
    \end{tabular}
  \end{center}

    \caption{WMT test set sizes. All test sets are filtered to use less than 128 tokens. This mainly reduced the en~\lfromto~de WMT'23 test set since this was a paragraph level task. The effect on all other test sets is minimal.}
    \label{tab:testset_filtering}
\end{table}

\section{Additional Results}
\label{sec:appendix-results}

\subsection{WMT Data: Domain-specific Evaluation}
\label{sec:wmt-domain-specific}

The WMT'22 and WMT'23 test sets are comprised of text originating from different domains.
The scores reported in the main text correspond to the evaluation of the corpora as a whole.
In Tables~\ref{tab:ende_domains} to~\ref{tab:jaen_domains} we show the results for each individual domain.
It can be seen that the improvements are achieved over all separate domains.
There are only two cases where training on all data performs slightly better than filtering with \bleurtqe
(ecommerce de~\lto~en in Table~\ref{tab:deen_domains}, and manuals zh~\lto~en in Table~\ref{tab:zhen_domains}).

\subsection{Other Metrics}
\label{sec:other-metrics}

In this appendix we report the \comet, \bleurt, \bleu and \chrF scores for the experiments reported in Section~\ref{sec:experiments}.
Table~\ref{tab:deen_comet_bleurt_bleu_chrf} shows the results for German \lto English, Table~\ref{tab:ende_comet_bleurt_bleu_chrf} for English \lto German, Table~\ref{tab:enja_comet_bleurt_bleu_chrf} for English \lto Japanese, Table~\ref{tab:jaen_comet_bleurt_bleu_chrf} for Japanese \lto English, Table~\ref{tab:enzh_comet_bleurt_bleu_chrf} for English \lto Chinese and Table~\ref{tab:zhen_comet_bleurt_bleu_chrf} for Chinese \lto English.
The additional metrics support the conclusions of the paper.

\renewcommand{\metricname}[1]{\textsc{\scriptsize #1}}
\renewcommand{\comet}{\textsc{Comet22}\xspace}

\begin{table*}[p]
  \begin{center}
    \small
    \begin{tabular}{lrrrrrrrr}
      \toprule
& \multicolumn{4}{c}{WMT'22} & \multicolumn{4}{c}{WMT'23} \\
        \cmidrule(lr){2-5} \cmidrule(lr){6-9}
& conversation & ecommerce & news & social & mastodon & news & speech & user review \\
      \midrule
	    \none      & \rp{87.93} & \rp{87.38} & \rp{86.37} & \rp{83.08} & \rp{82.47} & \rp{80.78} & \rp{80.87} & \rp{82.01} \\
	    \bicleaner & \rp{88.72} & \rp{88.06} & \rp{86.82} & \rp{83.14} & \rp{82.52} & \rp{82.28} & \rp{81.42} & \rp{82.10} \\
	    \cometqe   & \rp{88.75} & \rp{88.31} & \rp{86.76} & \rp{83.90} & \rp{83.25} & \rp{82.91} & \rp{81.57} & \rp{83.82} \\
	    \bleurtqe  & \textbf{\rp{88.92}} & \textbf{\rp{88.52}} & \textbf{\rp{87.18}} & \textbf{\rp{84.17}} & \textbf{\rp{84.18}} & \textbf{\rp{83.12}} & \textbf{\rp{81.77}} & \textbf{\rp{83.98}} \\
      \bottomrule
    \end{tabular}
  \end{center}
	\caption{\comet scores for each domain of en \lto de WMT test sets.}
    \label{tab:ende_domains}
\end{table*}

\begin{table*}[p]
  \begin{center}
    \small
    \begin{tabular}{lrrrrrrr}
      \toprule
& \multicolumn{4}{c}{WMT'22} \\
        \cmidrule(lr){2-5}
& conversation & ecommerce & news & social \\
      \midrule
	    \none      & \rp{84.73} & \textbf{\rp{85.40}} & \rp{84.41} & \rp{83.49} \\
	    \bicleaner & \rp{84.75} & \rp{84.96} & \rp{84.49} & \rp{82.80} \\
	    \cometqe   & \rp{84.96} & \rp{85.20} & \rp{84.82} & \rp{83.64} \\
	    \bleurtqe  & \textbf{\rp{85.11}} & \rp{85.31} & \textbf{\rp{84.89}} & \textbf{\rp{83.85}} \\
      \bottomrule
    \end{tabular}
  \end{center}
	\caption{\comet scores for each domain of de \lto en WMT test sets.}
    \label{tab:deen_domains}
\end{table*}

\begin{table*}[p]
  \begin{center}
    \small
    \begin{tabular}{lrrrrrrr}
      \toprule
& \multicolumn{4}{c}{WMT'22} \\
        \cmidrule(lr){2-5}
& conversation & ecommerce & news & social \\
      \midrule
\none      & \rp{84.19} & \rp{84.02} & \rp{81.52} & \rp{75.28} \\
\bicleaner & \rp{85.17} & \rp{83.86} & \rp{81.79} & \rp{76.11} \\
\comet     & \rp{86.03} & \rp{84.75} & \rp{83.29} & \rp{78.06} \\
	    \bleurt    & \textbf{\rp{86.38}} & \textbf{\rp{84.93}} & \textbf{\rp{83.61}} & \textbf{\rp{78.65}} \\
      \bottomrule
    \end{tabular}
  \end{center}
	\caption{\comet scores for each domain of en \lto zh WMT test sets.}
    \label{tab:enzh_domains}
\end{table*}

\begin{table*}[p]
  \begin{center}
    \small
    \begin{tabular}{lrrrrrrr}
      \toprule
& \multicolumn{4}{c}{WMT'22} & \multicolumn{3}{c}{WMT'23} \\
        \cmidrule(lr){2-5} \cmidrule(lr){6-8}
& conversation & ecommerce & news & social & manuals & news & user review \\
      \midrule
	    \none      & \rp{73.99} & \rp{66.80} & \rp{76.75} & \rp{74.13} & \textbf{\rp{77.67}} & \rp{78.92} & \rp{68.35} \\
	    \bicleaner & \rp{75.16} & \rp{70.81} & \rp{77.93} & \rp{75.62} & \rp{77.36} & \rp{79.50} & \rp{70.50} \\
	    \cometqe   & \textbf{\rp{76.39}} & \rp{71.19} & \rp{78.16} & \rp{75.73} & \rp{77.49} & \textbf{\rp{80.07}} & \rp{70.63} \\
	    \bleurtqe  & \rp{75.91} & \textbf{\rp{71.52}} & \textbf{\rp{78.30}} & \textbf{\rp{76.03}} & \rp{77.52} & \rp{79.67} & \textbf{\rp{71.03}} \\
      \bottomrule
    \end{tabular}
  \end{center}
	\caption{\comet scores for each domain of zh \lto en WMT test sets.}
    \label{tab:zhen_domains}
\end{table*}

\begin{table*}[p]
  \begin{center}
    \small
    \begin{tabular}{lrrrrrrr}
      \toprule
& \multicolumn{4}{c}{WMT'22} \\
        \cmidrule(lr){2-5}
& conversation & ecommerce & news & social \\
      \midrule
\none      & \rp{88.11} & \rp{87.48} & \rp{86.38} & \rp{80.61} \\
\bicleaner & \rp{88.60} & \rp{87.72} & \rp{86.54} & \rp{81.02} \\
\cometqe   & \rp{89.22} & \rp{87.91} & \rp{87.30} & \rp{81.91} \\
	    \bleurtqe  & \textbf{\rp{89.27}} & \textbf{\rp{88.57}} & \textbf{\rp{87.68}} & \textbf{\rp{82.52}} \\
      \bottomrule
    \end{tabular}
  \end{center}
	\caption{\comet scores for each domain of en \lto ja WMT test sets.}
    \label{tab:enja_domains}
\end{table*}

\begin{table*}[p]
  \begin{center}
    \small
    \begin{tabular}{lrrrrrrr}
      \toprule
& \multicolumn{4}{c}{WMT'22} \\
        \cmidrule(lr){2-5}
& conversation & ecommerce & news & social \\
      \midrule

	    \none      & \rp{77.46} & \rp{82.96} & \rp{75.20} & \rp{74.87} \\
	    \bicleaner & \rp{77.00} & \rp{83.14} & \rp{77.11} & \rp{75.23} \\
	    \cometqe   & \rp{77.38} & \textbf{\rp{84.14}} & \rp{77.84} & \rp{75.37} \\
	    \bleurtqe  & \textbf{\rp{78.21}} & \rp{83.90} & \textbf{\rp{78.32}} & \textbf{\rp{75.46}} \\

      \bottomrule
    \end{tabular}
  \end{center}
	\caption{\comet scores for each domain of ja \lto en WMT test sets.}
    \label{tab:jaen_domains}
\end{table*}

\renewcommand{\metricname}[1]{\textsc{#1}}
\renewcommand{\comet}{\metricname{Comet22}\xspace}

\renewcommand{\metricname}[1]{\textsc{\scriptsize #1}}

\begin{table*}
  \begin{center}
    \small
    \begin{tabular}{lrrrrrrrrrrrr}
      \toprule
       & \multicolumn{4}{c}{WMT'22 (dev)} & \multicolumn{4}{c}{WMT'19}   & \multicolumn{4}{c}{WMT'23}   \\
        \cmidrule(lr){2-5} \cmidrule(lr){6-9} \cmidrule(lr){10-13}
Filter     &        \comet & \bleurt       &         \bleu & \chrF         &        \comet & \bleurt       &         \bleu & \chrF         &        \comet & \bleurt       &         \bleu & \chrF         \\
\midrule
\none      &          84.5 & 73.4          &          32.3 & 57.4          &          84.6 & 73.1          &          41.1 & 64.8          &          83.3 & 72.2          &          37.4 & 61.2          \\
\random    &          84.2 & 73.1          &          32.0 & 57.1          &          84.3 & 72.9          &          40.5 & 64.4          &          83.5 & 72.2          &          38.2 & 62.1          \\
\bicleaner &          84.2 & 73.0          &          32.0 & 57.2          &          84.8 & 73.4          &          41.3 & 65.5          &          84.0 & 73.1          &          38.9 & 63.5          \\
\cometqe   &          84.6 & 73.6          & \textbf{32.5} & \textbf{57.5} &          85.1 & \textbf{74.0} & \textbf{41.8} & \textbf{65.7} & \textbf{84.6} & 73.8          & \textbf{40.6} & \textbf{65.0} \\
\bleurtqe  & \textbf{84.8} & \textbf{73.7} &          32.3 & 57.4          & \textbf{85.2} & \textbf{74.0} &          41.4 & 65.3          & \textbf{84.6} & \textbf{73.9} &          39.9 & 64.2          \\

\midrule

   & \multicolumn{4}{c}{IWSLT'21}   & \multicolumn{4}{c}{IWSLT'23}   \\
        \cmidrule(lr){2-5} \cmidrule(lr){6-9}
Filter     & \comet        & \bleurt       & \bleu         & \chrF         & \comet        & \bleurt       & \bleu         & \chrF         \\
\cmidrule[\lightrulewidth]{1-9}
\none      & 84.2          & 73.2          & 27.8          & 52.7          & \textbf{87.4} & \textbf{79.1} & \textbf{47.2} & \textbf{72.4} \\
\random    & 84.1          & 73.0          & 27.3          & 52.5          & 87.2          & 78.5          & 45.6          & 71.2          \\
\bicleaner & 84.1          & 73.0          & 27.5          & 52.5          & 87.3          & 79.0          & 46.9          & 72.1          \\
\cometqe   & \textbf{84.4} & \textbf{73.3} & \textbf{28.0} & \textbf{53.0} & 87.3          & 79.0          & 46.6          & 71.9          \\
\bleurtqe  & \textbf{84.4} & \textbf{73.3} & \textbf{28.0} & \textbf{53.0} & 87.3          & 79.0          & 47.1          & 72.2          \\

      \cmidrule[\heavyrulewidth]{1-9}  %
    \end{tabular}
  \end{center}

  \caption{Full results for German \lto English.
	The \cometqe and \bleurtqe results for IWSLT'21 are identical due to rounding.}
    \label{tab:deen_comet_bleurt_bleu_chrf}
\end{table*}

\begin{table*}
  \begin{center}
	\small
    \begin{tabular}{lrrrrrrrrrrrrrrrrrrrr}
      \toprule
                &  \multicolumn{4}{c}{WMT'22 (dev)}   & \multicolumn{4}{c}{WMT'19}   & \multicolumn{4}{c}{WMT'23}   \\
        \cmidrule(lr){2-5} \cmidrule(lr){6-9} \cmidrule(lr){10-13}
Filter     & \comet        & \bleurt       & \bleu         & \chrF         & \comet        & \bleurt       & \bleu         & \chrF         & \comet        & \bleurt       & \bleu         & \chrF         \\
\midrule
\none      & 86.2          & 76.7          & 35.9          & 62.6          & 86.0          & 75.7          & 43.1          & 67.0          & 81.8          & 70.1          & 38.2          & 61.6          \\
\random    & 86.0          & 76.5          & 35.1          & 62.2          & 85.5          & 75.0          & 42.4          & 66.5          & 80.8          & 68.9          & 36.4          & 61.0          \\
\bicleaner & 86.7          & 77.5          & 36.4          & 63.1          & 86.4          & 76.2          & 42.9          & 67.2          & 82.2          & 70.5          & 38.4          & 62.3          \\
\cometqe   & 86.9          & 77.6          & 36.2          & 63.0          & \textbf{86.7} & 76.4          & \textbf{44.0} & \textbf{68.0} & 82.9          & 71.9          & \textbf{40.9} & \textbf{65.9} \\
\bleurtqe  & \textbf{87.2} & \textbf{78.0} & \textbf{36.7} & \textbf{63.3} & \textbf{86.7} & \textbf{76.5} & 42.1          & 66.8          & \textbf{83.5} & \textbf{72.4} & \textbf{40.9} & 65.5          \\
\midrule
                  & \multicolumn{4}{c}{IWSLT'21}   & \multicolumn{4}{c}{IWSLT'23}   \\
        \cmidrule(lr){2-5} \cmidrule(lr){6-9} \\
Filter     & \comet        & \bleurt       & \bleu         & \chrF         & \comet        & \bleurt       & \bleu         & \chrF         \\
\cmidrule[\lightrulewidth]{1-9}
\none      & 83.2          & 73.0          & 23.2          & 56.6          & 84.5          & 76.5          & 45.8          & 72.2          \\
\random    & 82.8          & 72.8          & 22.8          & 56.3          & 84.2          & 76.2          & 44.9          & 71.7          \\
\bicleaner & 83.3          & 73.1          & 23.1          & 56.7          & 84.9          & \textbf{76.8} & 45.2          & 71.9          \\
\cometqe   & 83.6          & 73.5          & 23.2          & 56.9          & 84.8          & 76.6          & \textbf{45.9} & \textbf{72.4} \\
\bleurtqe  & \textbf{83.9} & \textbf{74.0} & \textbf{23.9} & \textbf{57.2} & \textbf{85.1} & \textbf{76.8} & 45.6          & 72.0          \\
      \cmidrule[\heavyrulewidth]{1-9}  %
    \end{tabular}
  \end{center}

    \caption{Full results for English \lto German.}
    \label{tab:ende_comet_bleurt_bleu_chrf}
\end{table*}

\begin{table*}
  \begin{center}
    \small
    \begin{tabular}{lrrrrrrrrrrrr}
      \toprule
                & \multicolumn{4}{c}{WMT'22 (dev)}   & \multicolumn{4}{c}{WMT'23}   & \multicolumn{4}{c}{IWSLT'23}   \\
        \cmidrule(lr){2-5} \cmidrule(lr){6-9} \cmidrule(lr){10-13}
Filter     & \comet        & \bleurt       & \bleu         & \chrF         & \comet        & \bleurt       & \bleu         & \chrF         & \comet        & \bleurt       & \bleu         & \chrF         \\
\midrule
\none      & 85.6          & 64.5          & 22.2          & 31.4          & 82.3          & 58.3          & 19.0          & 28.9          & 86.9          & 68.1          & 40.5          & 48.1          \\
\random    & 84.5          & 62.8          & 21.0          & 30.3          & 80.7          & 55.5          & 17.2          & 27.2          & 85.8          & 65.8          & 35.7          & 43.6          \\
\bicleaner & 86.0          & 64.9          & 21.9          & 31.5          & 83.2          & 59.2          & 19.1          & 29.0          & 87.2          & 68.2          & 39.2          & 47.1          \\
\cometqe   & 86.6          & 65.5          & \textbf{22.7} & 32.1          & 83.7          & \textbf{60.0} & 19.3          & \textbf{29.5} & \textbf{87.9} & \textbf{69.3} & \textbf{41.3} & \textbf{48.6} \\
\bleurtqe  & \textbf{87.0} & \textbf{66.1} & \textbf{22.7} & \textbf{32.2} & \textbf{84.0} & \textbf{60.0} & \textbf{19.5} & \textbf{29.5} & 87.4          & 68.1          & 38.8          & 46.5          \\
      \bottomrule
    \end{tabular}
  \end{center}

    \caption{Full results for English \lto Japanese.}
    \label{tab:enja_comet_bleurt_bleu_chrf}
\end{table*}

\begin{table*}
  \begin{center}
    \small
    \begin{tabular}{lrrrrrrrrrrrr}
      \toprule
                & \multicolumn{4}{c}{WMT'22 (dev)}   & \multicolumn{4}{c}{WMT'23}   & \multicolumn{4}{c}{IWSLT'23}   \\
        \cmidrule(lr){2-5} \cmidrule(lr){6-9} \cmidrule(lr){10-13}
 Filter    & \comet        & \bleurt       & \bleu         & \chrF         & \comet        & \bleurt       & \bleu         & \chrF         & \comet        & \bleurt       & \bleu         & \chrF         \\
\midrule
 \none     & 77.6          & 62.6          & 18.1          & 42.5          & 75.9          & 62.2          & 16.0          & 41.6          & \textbf{85.5} & \textbf{73.4} & \textbf{30.5} & \textbf{62.2} \\
\random    & 75.9          & 60.8          & 16.1          & 40.8          & 75.0          & 61.0          & 15.0          & 40.7          & 84.6          & 71.9          & 27.4          & 59.9          \\
\bicleaner & 78.1          & 63.6          & 18.2          & 44.3          & 77.4          & 63.3          & 17.2          & 44.2          & 85.0          & 72.2          & 28.4          & 61.0          \\
\cometqe   & 78.7          & 64.1          & 18.8          & 44.6          & 78.0          & 64.0          & 16.6          & 44.1          & 85.0          & 72.6          & 28.2          & 60.8          \\
\bleurtqe  & \textbf{79.0} & \textbf{64.4} & \textbf{19.0} & \textbf{45.2} & \textbf{78.2} & \textbf{64.3} & \textbf{17.4} & \textbf{44.7} & 85.1          & 72.8          & 29.6          & 61.7          \\
      \bottomrule
    \end{tabular}
  \end{center}

    \caption{Full results for Japanese \lto English.}
    \label{tab:jaen_comet_bleurt_bleu_chrf}
\end{table*}

\begin{table*}
  \begin{center}
	\small
    \begin{tabular}{lrrrrrrrr}
      \toprule
                & \multicolumn{4}{c}{WMT'22 (dev)}   & \multicolumn{4}{c}{WMT'23}   \\
        \cmidrule(lr){2-5} \cmidrule(lr){6-9}
           & \comet        & \bleurt       & \bleu         & \chrF         & \comet        & \bleurt       & \bleu         & \chrF         \\
\midrule
\none      & 81.2          & 65.8          & 37.7          & 33.5          & 79.7          & 64.8          & 43.4          & 39.0          \\
\random    & 80.2          & 64.5          & 36.6          & 32.7          & 79.3          & 64.3          & 41.6          & 37.3          \\
\bicleaner & 81.7          & 66.6          & 37.0          & 33.0          & 80.3          & 65.7          & 42.4          & 37.6          \\
\cometqe   & 83.0          & 68.0          & 38.2          & 34.0          & \textbf{82.5} & \textbf{68.2} & \textbf{44.0} & \textbf{40.1} \\
\bleurtqe  & \textbf{83.4} & \textbf{68.5} & \textbf{38.7} & \textbf{34.4} & 82.2          & 67.7          & 43.6          & 38.9          \\
\midrule
           & \multicolumn{4}{c}{WMT'19} & \multicolumn{4}{c}{IWSLT'23} \\
        \cmidrule(lr){2-5} \cmidrule(lr){6-9}
           & \comet                     & \bleurt                       & \bleu         & \chrF         & \comet        & \bleurt       & \bleu         & \chrF         \\
\midrule
\none      & 77.6                       & 59.4                          & 31.1          & 27.4          & \textbf{84.2} & \textbf{72.2} & \textbf{52.5} & 47.1          \\
\random    & 77.2                       & 59.1                          & 30.7          & 27.5          & 82.1          & 69.3          & 47.6          & 41.9          \\
\bicleaner & 78.4                       & 60.4                          & 31.1          & 27.5          & 83.3          & 70.7          & 47.9          & \textbf{42.3} \\
\cometqe   & \textbf{80.1}              & \textbf{62.2}                 & \textbf{32.1} & \textbf{28.3} & 84.1          & 71.2          & 47.9          & 42.2          \\
\bleurtqe  & 79.9                       & 61.9                          & 31.7          & 28.1          & 84.1          & 71.1          & 47.3          & 42.2          \\
      \bottomrule
    \end{tabular}
  \end{center}

    \caption{Full results for English \lto Chinese.}
    \label{tab:enzh_comet_bleurt_bleu_chrf}
\end{table*}

\begin{table*}
  \begin{center}
	\small
    \begin{tabular}{lrrrrrrrr}
      \toprule
                   & \multicolumn{4}{c}{WMT'22 (dev)}   & \multicolumn{4}{c}{WMT'23}   \\
        \cmidrule(lr){2-5} \cmidrule(lr){6-9}
           & \comet        & \bleurt       & \bleu         & \chrF         & \comet        & \bleurt       & \bleu         & \chrF         \\
\midrule
\none      & 72.8          & 59.5          & 17.4          & 46.2          & 74.7          & 61.0          & 18.8          & 44.9          \\
\random    & 72.2          & 58.8          & 16.9          & 46.3          & 74.2          & 60.2          & 18.5          & 44.9          \\
\bicleaner & 74.8          & 60.9          & 17.1          & 47.6          & 75.7          & 61.5          & 19.1          & 45.8          \\
\cometqe   & 75.2          & 61.4          & \textbf{17.9} & \textbf{48.5} & \textbf{76.0} & 61.6          & \textbf{19.2} & \textbf{46.2} \\
\bleurtqe  & \textbf{75.4} & \textbf{61.7} & 17.7          & 48.1          & \textbf{76.0} & \textbf{61.9} & 18.6          & 45.7          \\
\midrule
                   & \multicolumn{4}{c}{WMT'19}   & \multicolumn{4}{c}{IWSLT'23}   \\
        \cmidrule(lr){2-5} \cmidrule(lr){6-9}
           & \comet        & \bleurt       & \bleu         & \chrF         & \comet        & \bleurt       & \bleu         & \chrF         \\
\midrule
\none      & 78.3          & 65.6          & 23.7          & 53.6          & \textbf{84.9} & \textbf{74.5} & \textbf{33.3} & \textbf{63.2} \\
\random    & 78.1          & 65.0          & 23.2          & 53.2          & 84.1          & 73.8          & 31.7          & 62.1          \\
\bicleaner & 79.6          & 66.6          & 23.9          & 54.7          & 84.2          & 73.9          & 32.6          & 62.4          \\
\cometqe   & \textbf{80.0} & 67.0          & \textbf{24.7} & \textbf{55.5} & 84.5          & 73.9          & 30.8          & 61.6          \\
\bleurtqe  & 79.9          & \textbf{67.2} & 23.7          & 54.9          & 84.8          & \textbf{74.5} & 32.1          & 62.5          \\
      \bottomrule
    \end{tabular}
  \end{center}

    \caption{Full results for Chinese \lto English.}
    \label{tab:zhen_comet_bleurt_bleu_chrf}
\end{table*}

\end{document}